\pdfoutput=1
\documentclass[11pt]{article}

\usepackage[]{EMNLP2022}

\usepackage{times}
\usepackage{latexsym}

\usepackage[T1]{fontenc}
\usepackage[utf8]{inputenc}

\usepackage{microtype}

\usepackage{inconsolata}

\usepackage{cleveref}
\usepackage{multirow}
\usepackage{arydshln}
\usepackage{graphicx}
\graphicspath{ {./figures/} }

\title{Knowledge Prompts: Injecting World Knowledge into \\ Language Models through Soft Prompts}

\author{Cicero Nogueira dos Santos, Zhe Dong, Daniel Cer, John Nham, \\ 
\textbf{Siamak Shakeri, Jianmo Ni, Yun-hsuan Sung} \\
  Google Research \\
  \texttt{$\{$cicerons, zhedong, cer, jnham, siamaks, jianmon, yhsung$\}$@google.com} \\}
  
\begin{document}
\maketitle
\begin{abstract}
Soft prompts have been recently proposed as a tool for adapting large frozen language models (LMs) to new tasks.
In this work, we repurpose soft prompts to the task of injecting world knowledge into LMs.
We introduce a method to train soft prompts via self-supervised learning on data from knowledge bases.
The resulting soft knowledge prompts (KPs) are task independent and work as an external memory of the LMs. 
We perform qualitative and quantitative experiments and demonstrate that: (1) KPs can effectively model the structure of the training data; (2) KPs can be used to improve the performance of LMs in different knowledge intensive tasks.
\end{abstract}

\section{Introduction}
Very large neural language models (LMs) are known to perform well on knowledge intensive natural language understanding (NLU) tasks, because they memorize a significant amount of world knowledge from the training data. 
The larger the LM, the more facts it can memorize at the training time, and the better the results at the inference time \cite{roberts-etal-2020-much}.
Despite their success, these models also present some important drawbacks such as:
the parametric memory of these models have a fixed size and cannot grow (or shrink) over time without fully retraining the model;
there is no control in terms of which part of the memory stores data about what;
facts that do not co-occur frequently in the training data are not well represented in the model; 
very large models are needed to memorize enough data in order to perform well on knowledge intensive tasks such as generative question answering; and at last, but not the least, the memorized knowledge gets obsolete overtime, and requires re-training the model for refreshness.

In this work, we employ soft prompts to overcome some of these issues of LMs.
\emph{Soft prompts} \cite{lester-etal-2021-power,li-liang-2021-prefix,hambardzumyan-etal-2021-warp}  have been recently proposed as a tool for adapting large frozen LMs to new tasks.
Nevertheless, we repurpose soft prompts to the task of injecting world knowledge into LMs.
The goal is to train an external memory that is composed of a large set of soft prompts that encode world knowledge.
We introduce a method to train knowledge driven soft prompts via self-supervised learning on data from knowledge bases.
The resulting soft prompts, which we call \emph{knowledge prompts} (KPs), function as an auxiliary memory of the LMs that is activated when solving knowledge intensive tasks.
Different from regular applications of soft prompts that concatenate a fixed small set of embeddings to every input,
our approach learns a very large set of KPs,
which are sparsely activated depending on the input.

We focus on entity-centric KPs, which means that each prompt primarily encodes information about one entity from a knowledge base.
We use Wikidata \cite{wikidata_2014} triples as our training data and train KPs for the top 1.1M entities, based on the number of triples.
We present a qualitative analysis of KPs using t-SNE plots and k-nearest neighbors approaches.
In terms of quantitative analysis, 
we show experimental results for three knowledge intensive tasks: question answering, fact checking and relation classification.
For all datasets, the use of KPs improves the performance of the T5 baseline.
Our experimental results demonstrate that KPs are an effective way to expand the memory of frozen LMs.

%to inject world knowledge into LMs in a way that is useful to improve the performance on downstream tasks.

%Our experimental results demonstrate that KPs are an effective way to inject world knowledge into LMs in a way that is useful to improve the performance on downstream tasks.

\begin{figure*}[ht!]
    \centering
    \includegraphics[scale=.38]{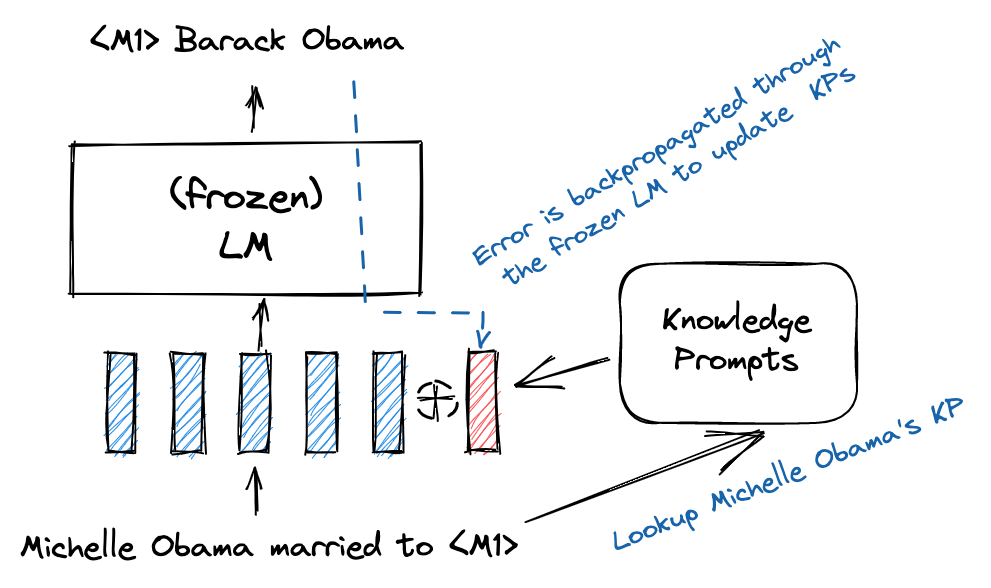}
    \caption{\textbf{Training of Knowledge Prompts}: given a serialized KB triple where one of the entities has been masked out, the frozen LM has to predict the masked entity given the input and the knowledge prompt of the non-masked entity, which is \emph{Michelle Obama} in the example. The cross-entropy loss is computed and the error is back-propagated through the frozen LM in order to update the KP.}
    \label{fig:kp-training}
\end{figure*}

The main contributions of this work are the following:
\begin{itemize}
    \item we propose a self-supervised approach to train knowledge driven soft prompts that can be used to inject world knowledge into LMs.
    \item we demonstrate that knowledge prompts can effectively model the structure of the training data and can also improve the performance of LMs on knowledge intensive tasks.
    \item this work sheds light on the usability of soft prompts for storing data rather than storing instructions on how to solve specific tasks.
\end{itemize}

\section{Methods}

\subsection{Soft Prompts}
Different approaches have been recently proposed to train soft prompts \cite{lester-etal-2021-power,li-liang-2021-prefix,hambardzumyan-etal-2021-warp}. 
One of the most popular methods, and probably the simplest one, consists of the following steps \citep{lester-etal-2021-power}: 
\begin{itemize}
    \item[(1)] for a task in the dataset, prepend a fixed number of embeddings (soft prompts) to the word embeddings of every input;
    \item[(2)] during finetuning, update the soft prompt while keeping all the other parameters of the LM frozen. 
\end{itemize}
Despite its simplicity, this approach has demonstrated to be very effective when used with large language models.

\subsection{Soft Knowledge Prompts}
We are interested in training \emph{soft knowledge prompts} (KPs) to encode world knowledge, which could work as an external memory for LMs.  
% Our goal is to train a large set of soft prompts that works as an external memory of the LM.
% Therefore, instead of training soft prompts to solve individual tasks, we are interested in training \emph{soft knowledge prompts} (KPs) that encode world knowledge.
In this work, we focus on the training of \emph{entity-centric KPs}, 
% as it gives us a clear way to decide which pieces of knowledge should be stored in each soft prompt.
each of which stores the knowledge related to a specific entity from a knowledge base (KB). 
In other words, the KP of an entity encodes information from the KB triples that mention the entity either as a subject or an object. 
% In our training strategy, 
We adopt KB triples from Wikidata \cite{wikidata_2014}, as a simple and trustworthy source of world knowledge.
% In our setting, each KP stores information about one entity from the knowledge base. 
% In other words,
% the KP of an entity encodes information from the KB triples that mention that entity either as a subject or object. 

\subsubsection{KP Training}
We train KPs with a masked language modeling (MLM) objective \citep{devlin-etal-2019-bert, Taylor1953ClozePA},
% Our method to train KPs uses a masked language modeling (MLM) objective
where the goal is to generate the object entity of a KB triple given the subject entity and relation, and vice versa.
As an example, the input / target pair "\texttt{Germany capital <MASK>}" / "\texttt{Berlin}" will be used to update the KP for \texttt{Germany}, while the pair "\texttt{<MASK> capital Berlin}" / "\texttt{Germany}" will be used to update the KP for \texttt{Berlin}.

The KPs are randomly initialized,
and are updated only when the corresponding entities appear (not masked) in the input.  This makes the training of KPs sparse and parallelizable.

Given an input triple with the object entity being masked, a training iteration has the following steps:
\begin{itemize}
    \item[(1)] retrieve the KP of the subject entity, which is a simple lookup operation;
    % since the entity is known;
    \item[(2)] concatenate the KP to the sequence of word embeddings of the input text;
    \item[(3)] predict the object entity name and compute the cross-entropy loss;
    \item[(4)] back-propagate the error through the frozen LM to the KP, and update the KP using stochastic gradient descent;
\end{itemize}
\Cref{fig:kp-training} illustrates our proposed method for training knowledge prompts.
Notice that this method is general, and can be used with any textual input as long as the entities of interest are already identified/linked.

%In our particular case, we can think of the resulting KPs as an entity embedding matrix.

\subsubsection{Using KPs in Downstream Tasks}
Using KPs during the finetuning of the LM is straightforward. Given the input sequence, e.g. a question, the relevant KPs are retrieved and concatenated to the word embeddings of the input, to generate or predict the answer.
At the finetuning time,
% Since our KP training approach produces general KPs, 
we freeze the KPs and only update the parameters of the LM, as the KPs are  used as pre-trained external knowledge.
One can also use task specific soft prompts instead of finetuning the parameters of the LM as in \cite{lester-etal-2021-power}, however in this work we focused on finetuning the LM as we use relatively small LMs.

Retrieving KPs that are relevant to the input sequence is crucial for good performance in the downstream task. KPs are useful only if they contain the knowledge that is helpful to solve the input at hand.
In this work we employ entity linking as a way to retrieve relevant KPs during training/inference for downstream tasks.
Given an input, we first perform entity linking to identify the entities mentioned in the input. Then, we simply do a lookup operation to retrieve the KPs of the identified entities.

\subsubsection{Why Knowledge Prompts?}
Some advantages of the proposed approach include:
\begin{itemize}
    \item It allows a better control of what information is stored, by choosing what examples are used to train the KPs.
    \item KPs are trained independently, therefore the training can be massively parallelized.
    \item As the LM is kept frozen during the training of KPs, we do not mess up with the language generation/understanding capabilities of the LM.
    \item KPs can increase the capacity of small LMs in a dynamic way. We can add/remove KPs at any time. Moreover, if information about a single entity changes, we can update that entity's KP without changing other KPs. This addresses the freshness issue of the LMs.
\end{itemize}
\section{Related Work}
Most works on soft prompting focus on the problem of learning a set of embeddings that can \emph{reprogram} a frozen LM to solve a specific task \cite{lester-etal-2021-power,li-liang-2021-prefix,hambardzumyan-etal-2021-warp}.
Some key aspects that distinct KPs from these approaches include:
(1) in KPs, the goal is to create an external memory that stores world knowledge.
(2) while traditional soft prompt approaches learn a small set of embeddings our approach learns a very large set of KPs, which are sparsely accessed by the LM. 
(3) different from regular soft prompts, KPs are not task-specific.

%,vu-etal-2022-spot

Our work is related to the recent body of work on memory augmented LMs.
The most related approaches are the Entities as Experts (EaE) model proposed by \cite{fevry-etal-2020-entities}
and the Fact Injected Language Model (FILM) proposed by \cite{verga-etal-2021-adaptable}.
In both papers the authors present methods to train entity embeddings during the pretraining of the LM.
During inference time, the model uses a separate classification head to identify entities, which are then used to retrieve entity embeddings that are merged to the current entity representation.
\cite{jong2022mention} proposed the TOME model,
which employs embeddings in a similar way to EaE but with a much larger granularity.
Instead of modeling entities, their method embeds entity mentions, which results in a very large external memory. 
In their experiments, they use up to 150M entries in the external memory while EaE, FILM and our method use about 1M entries only.
Compared to these three models, 
our proposed approach trains an external LM memory in a quite different way. 
Our method uses frozen pretrained LMs while those methods train the external memory and the LMs together.
Our method concatenates the relevant memory entry to the input as if they were additional word embeddings, while the other three approaches merge the relevant memory entries to the contextual embedding of the identified entities. 
Additionally, while EaE, FILM and TOME adapt the memory for each new task, our approach uses the exact same external memory for different tasks.

This work is also related to multiple recent papers on integrating knowledge bases into LMs
\cite{zhang-etal-2019-ernie,peters-etal-2019-knowbert,poerner-etal-2020-e,sun-etal-2020-colake,wang-etal-2021-kepler,agarwal-etal-2021-knowledge}.
On of the key differences between KPs and all these methods is the use of soft prompts to integrate knowledge bases to frozen LMs.
\section{Experimental Setup}

%\noindent{\textbf{KP Training Data:}}
\subsection{KP Training Data}
We adopt Wikidata triples \cite{wikidata_2014} as our source of data to train KPs.
We start with the set of 45M triples that was previously preprocessed by \cite{agarwal-etal-2021-knowledge}.
Next, we filter out triples whose subject entity appears less than 12 times as subject entity in the dataset.
This results in a set of 23M triples containing 1.1M distinct subject entities, which form our entity vocabulary and, respectively, the number of KPs in our experiments.

%\noindent{\textbf{KP Training setup:}} 
\subsection{KP Training Setup}
We adopt the T5.1.1 model family \cite{2020t5} and perform experiments with three model sizes: small, base and large,
which contain 60M, 220M and 770M parameters, respectively.
We use the T5.1.1 checkpoints that  \citet{lester-etal-2021-power} adapted from the original T5.1.1 checkpoints by running an additional 100K training steps using the “LM” objective discussed in \cite{2020t5}.
Just like reported by \citet{lester-etal-2021-power}, we also noticed that these adapted checkpoints make the training of soft prompts easier. 
Although we use an encoder-decoder LM, our approach is not limited to this type of architecture and can be used with encoder only models like BERT \cite{devlin-etal-2019-bert} or decoder only models like GPT2 \cite{radford2019language}.

The input length for training KPs is normally short because our examples are masked serialized triples (concatenation of Subject/Object entity and a relation). Therefore, we set the input length to 64, which allows us to use very large batch sizes: between 4K and 8K, depending on the model size. 
Note that the objective when training KPs is to memorize the training data. Hence, we let KP training run for up to 200 epochs.

In the beginning of the training, KPs are initialized by randomly sampling from the word embeddings matrix.
This allows KPs to start from a region that is known by the LM, which makes the training smoother and less sensitive to hyperparameters.
After training, KPs are kept frozen during LM finetuning for downstream tasks.
Therefore, for each model size, the exact same set of KPs is used in our experiments with the different downstream tasks and datasets. 

%\noindent{\textbf{Entity Linking:}} 
\subsection{Entity Linking}
In all experiments where KPs are used,
we first preprocess the input text using Google Cloud Natural Language API \footnote{https://cloud.google.com/natural-language/docs/analyzing-entities} to perform entity linking.

%\noindent{\textbf{Downstream Tasks:}} 
\subsection{Downstream Tasks}
We perform experiments with three different knowledge intensive tasks:  
(1) question answering (QA), 
(2) fact checking and 
(3) relation classification.
In terms of datasets, for question Answering experiments we use Entity Questions \cite{sciavolino-etal-2021-simple} and TriviaQA \cite{joshi-etal-2017-triviaqa} datasets. 
For fact checking we use the FEVER dataset \cite{thorne-etal-2018-fever}. 
For relation classification, we use the TACRED dataset \cite{zhang-etal-2017-position}.

In the question answering experiments we follow the closed-book QA (CBQA) setup of \citet{roberts-etal-2020-much}.
In this setup, the model has no access to external text, which means that there is no retrieval step and the model has to solve the task using the world knowledge it acquired from pretraining and finetuning data only.
During training, we try to use the default hyperparameters as much as possible except for the learning rate, 
which is finetuned on the development sets.
Following previous works we use exact matching (EM) as the evaluation metric in CBQA.

\section{Results and Discussion}

\begin{figure*}[ht!]
    \centering
    \includegraphics[scale=.13]{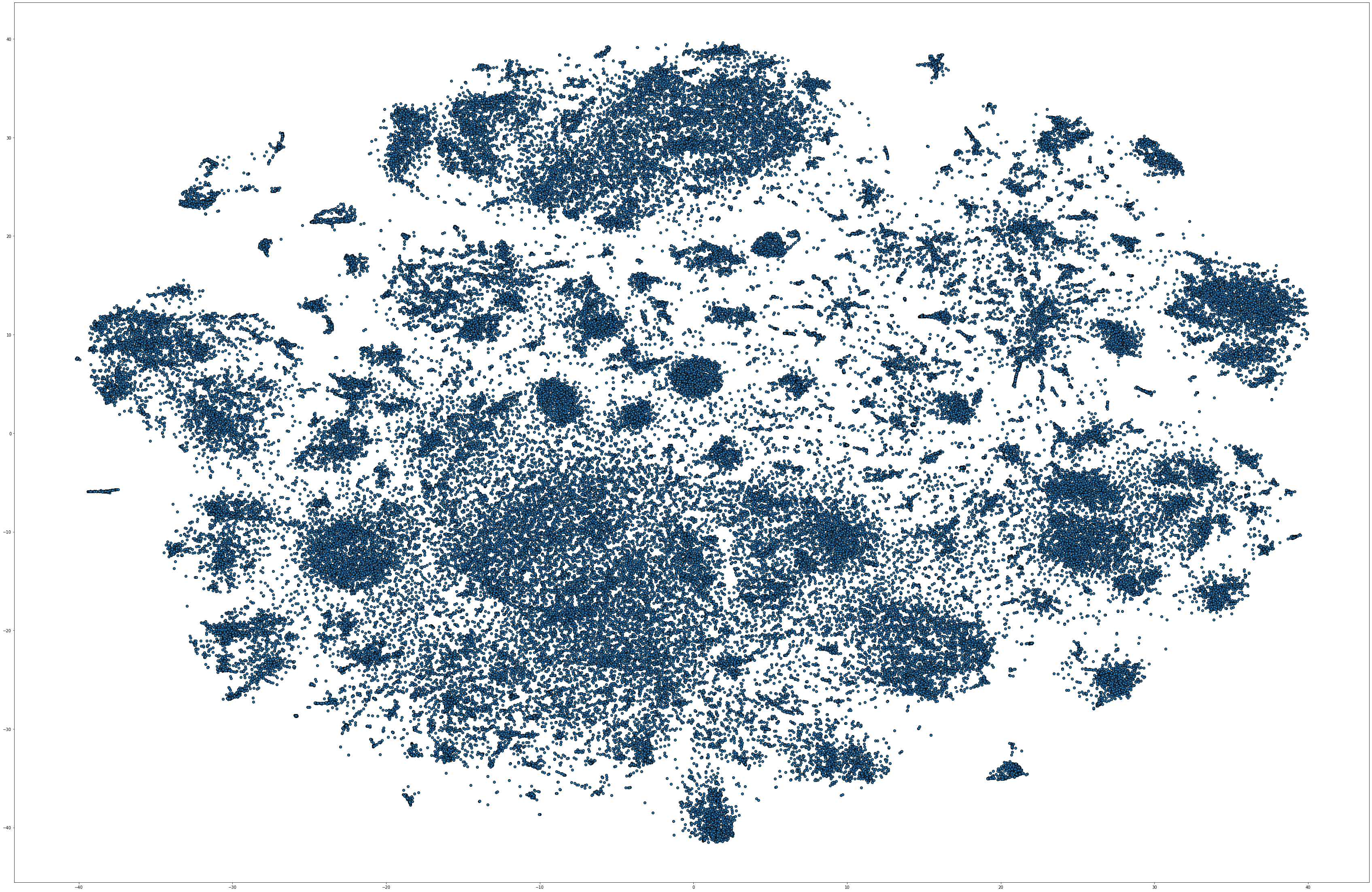}
    \caption{t-SNE visualization of 100K KPs trained with T5-Base model on Wikidata. KPs form well separated clusters whose member entities are very similar in terms of their properties and the relations they belong to.}.
    \label{fig:kp-tsne}
\end{figure*}

\subsection{Qualitative Results}
We perform a qualitative assessment of KPs through different experiments including t-SNE visualizations, analysis of entity similarity in KP space and evaluation of KPs for QA when golden entity linking is provided.

\subsubsection{t-SNE visualization of KPs}
One of the main goals in our qualitative assessment of KPs is to check whether the learned KPs can model the structure of the training data.
An approach that can give us some clue about the data structure learning aspect are t-SNE visualizations \cite{vandermaaten08a_tsne}.
In Fig. \ref{fig:kp-tsne}, we show a t-SNE visualization of 100K randomly selected KPs that were trained using T5-BASE model on Wikidata triples.
We can see in Fig. \ref{fig:kp-tsne} that KPs form multiple well separated clusters.
Zooming in into these clusters we can notice that they are very coherent. 
There are clusters about companies, celestial objects, movies, locations, etc.
This is a good indication that, although trained independently, KPs encapsulate a notion of similarity between entities that aligns well with the KB structure.

\subsubsection{k-Nearest Neighbors in KP Space}
We investigate further the quality of the entity similarity captured by KPs using cosine similarity in the KP space to retrieve the k-nearest neighbors of different entities.
In Table \ref{tab:results:knn} we show the top four neighbours of four different entities.
We present results for T5 models of three different sizes.
The top two entities (\emph{Barack Obama} and \emph{Roger Waters}) are cherry picked popular entities to make it easier for the reader to grasp the quality of the results.
The bottom two entities (\emph{Fairmont station} and \emph{Iacobeni , Sibiu}) were randomly selected.
The KP space learned by the three model sizes can produce high quality nearest neighbors for the four different entities.
For instance, in the case of the search entity \emph{Fairmont station}, which is a streetcar stop in Salt Lake City - Utah - USA, all the retrieve entities are also streetcar stops in Salt Lake City.
Similar results can be seen for the other entities, where the retrieved neighbors share multiple properties (e.g. same occupation and nationality) with the search entity.

\begin{table*}[ht!]
\centering
\begin{tabular}{lccc}
\hline
\textbf{Search Entity } & \multicolumn{3}{c}{\textbf{Top 4 neighbors in KP space}} \\
& \multicolumn{1}{c}{\textbf{Small}} & \multicolumn{1}{c}{\textbf{Base}} & \multicolumn{1}{c}{\textbf{Large}}  \\
\hline
\multirow{3}{*}{Barack Obama} & Donald Trump & Bill Clinton & Richard Nixon  \\
                              & George W. Bush & Donald Trump & Donald Trump \\
                              & Michelle Obama & Ronald Reagan & Michelle Obama  \\
                              & Jimmy Carter & Jimmy Carter & Harrison Ford  \\
                            %   & Franklin D. Roosevelt & Michelle Obama & Joe Biden  \\
\hdashline
\multirow{3}{*}{Roger Waters} & George Harrison & David Gilmour & Tom Waits  \\
                              & Pink Floyd & Syd Barrett & Freddie Mercury  \\
                              & Kris Kristofferson & Eddie Vedder & David Gilmour  \\
                              & James Hetfield & Brian Wilson & Cliff Edwards  \\
                            %   & Geffen Records & Pink Floyd & Keith Richards  \\
\hdashline
\multirow{3}{*}{Fairmont station} & 500 East station & 700 East station & 700 East station  \\
                              & 300 East station & 300 East station & Historic Gardner station  \\
                              & Sugarmont station & 500 East station & 300 East station  \\
                              & 700 East station & Fairpark station & Decker Lake station  \\
                            %   & 4800 West Old Bingham Highway station & 1940 West North Temple station & 2700 West Sugar Factory Road station  \\
\hdashline
\multirow{3}{*}{Iacobeni , Sibiu} & Pălatca & Mica , Cluj & Sic , Cluj  \\
                              & Racovița , Timiș & Gârbău , Cluj & Gârbău , Cluj  \\
                              & Movila , Ialomița & Aiton , Cluj & Mihăileni , Sibiu  \\
                              & Moșna , Sibiu & Râșca , Cluj & Mica , Cluj  \\
                            %   & Sagna , Neamț & Târnava , Sibiu & Jina , Sibiu  \\
\hline
\end{tabular}
\caption{Nearest neighbors in the knowledge prompt space for different search entities and model sizes. The top two entities were cherry picked while the two bottom ones were randomly selected.}
\label{tab:results:knn}
\end{table*}

\begin{table*}[ht!]
\centering
\begin{tabular}{lcccc}
\hline
\textbf{Model} & \multicolumn{4}{c}{\textbf{Simple Questions}}  \\
& \multicolumn{2}{c}{\textbf{Zero-shot Learning}} & \multicolumn{2}{c}{\textbf{Finetuning}} \\
& no KPs & KPs & no KPs & KPs \\
\hline
T5-SMALL & 0.1 & \textbf{4.3} & 32.9 & \textbf{54.1} \\
T5-BASE  & 0.0 & \textbf{8.8} & 35.1 & \textbf{58.9} \\
T5-LARGE & 0.0 & \textbf{8.8} & 36.6 & \textbf{58.3} \\
\hline
\end{tabular}
\caption{Comparing LMs performances on Simple Questions \cite{BordesUCW15_simple_questions} with and without KPs as a knowledge source, measured by exact match (EM) score (\%). The KPs are retrieved using golden entity linking information. Most performant results for each setup is marked in bold.}
\label{tab:results:simplequestions}
\end{table*}

\subsubsection{KPs as a Knowledge Source for LMs}
In order to assess in a controlled manner whether KPs can be used as a knowledge source for LMs, we perform an experiment on closed-book QA using the Simple Questions dataset \cite{BordesUCW15_simple_questions}.
This dataset is a good fit for our purpose because it contains golden information about the entity involved in the question (golden entity linking).
We use the Simple Questions version that was processed by \citet{wikidata-benchmark} to align the original Freebase entities to Wikidata entities.
We further preprocessed the dataset to contain only questions involving one of the 1.1M entities for which we trained KPs.

In Table \ref{tab:results:simplequestions},
we present two sets of experiments for models of different sizes.
In the first experiment we check whether the use of KPs can improve the performance of the models for zero-shot learning.
In this scenario,
we can see that T5 models without KPs performs very poorly and achieve exact match (EM) score of 0 percent.
The use of KPs boosts the performance of all model sizes, with the base and large models achieving EM of 8.8.
In the finetuning scenario,
the use of KPs also brings a significant boost in performance.
In particular, for T5-Base the improvement is of  $\sim$24 points in EM.
We know the improvement is actually larger than this because for some questions there are multiple good answers (e.g. songwriter and singer are valid occupations for \emph{John Lennon}),
but the Simple Questions dataset list a single answer only.

These experimental results indicate that KPs are an effective approach to store information about entities in a way that can be readily used by the LM without any adaptation of the KPs for the downstream QA task. 

\subsection{Quantitative Results}

\begin{table*}[t!]
\centering
\begin{tabular}{lcccc|cccc}
\hline
\textbf{Model} & \multicolumn{4}{c}{\textbf{Entity Questions}} & \multicolumn{4}{c}{\textbf{TriviaQA}} \\
& \multicolumn{2}{c}{\textbf{Dev}} 
& \multicolumn{2}{c}{\textbf{Test}} 
& \multicolumn{2}{c}{\textbf{Dev}} 
& \multicolumn{2}{c}{\textbf{Test}} \\
& no KPs & KPs & no KP & KPs & no KPs & KPs & no KP & KPs \\
\hline
T5-SMALL & 23.3 & \textbf{30.6} & 23.6 & \textbf{30.8} & 17.4 & \textbf{19.4} & 21.4 & \textbf{23.8} \\
T5-BASE  & 24.5 & \textbf{33.3} & 25.4 & \textbf{33.1} & 22.7 & \textbf{24.9} & 27.1 & \textbf{28.1}  \\
T5-LARGE & 26.8 & \textbf{33.8} & 26.7 & \textbf{33.9} & 28.3 & \textbf{28.9} & 32.0 & \textbf{32.5} \\
\hline
% BERT-base \cite{fevry-etal-2020-entities} &   &  &  &  & 28.9 &  &  &  \\
% T5-11B \cite{roberts-etal-2020-much} &   &  &  &  & 42.3  &  & 50.1 &  \\
% EaE \cite{fevry-etal-2020-entities} &   &  &  &  & 43.2  &  & 53.4 &  \\
% T5-11B + SSM \cite{roberts-etal-2020-much} &   &  &  &  & 51.0  &  & 60.5 &  \\
% TOME \cite{jong2022mention} &  &  &  &  & 54.6  &  & 65.8 &  \\
% \hline
\end{tabular}
\caption{Comparing LMs performances on EntityQuestions \citep{sciavolino-etal-2021-simple} and TriviaQA \citep{joshi-etal-2017-triviaqa} with and without KPs as a knowledge source, measured by exact match (EM) score (\%).}
\label{tab:results:qa}
\end{table*}

\subsubsection{Closed-book Question Answering}
Table \ref{tab:results:qa} presents experimental results for two closed-book QA (CBQA) datasets and different T5 model sizes.
KPs provide a significant performance improvement on the Entity Questions dataset, which is an entity-centric QA task.
For instance, the improvement for T5-Base is of 7.7 points in EM.
Interestingly,
T5-SMALL + KPs outperforms T5-LARGE model by a large margin, 30.8 vs 26.7, respectively.
Although to a smaller extent,
KPs also bring performance improvements for all model sizes on TriviaQA dataset.

In Table \ref{tab:results:qa_related} we compare the performance of T5-Base/Large + KPs with other recently published results on CBQA for TriviaQA.
To the best of our knowledge, there is no previous work that reports CBQA results for the Entity Questions dataset.
T5 + KPs model does not perform as well as the other CBQA approaches on TriviaQA.
We believe this is mainly due to the following factors:
(1) in EaE, TOME and T5-11B+SSM, the full LM is heavily pretrained on Wikipedia text using entity focused objective functions, which is known to make the model very effective to QA \cite{guu2020realm}.
In our method, 
we use an entity focused objective to train KPs only while the LM is frozen.
Note in Table \ref{tab:results:qa}, no KPs column, that our initial baseline is poor.
(2) Models like EaE and TOME update their external memory component during finetuning for a new task. We keep KPs frozen when finetuning for a new task.
(3) Our model generates the answer token by token, which is more prone to errors compared to the entity retrieval approaches used in EaE and TOME.
(4) We conjecture that the additional pretraining steps performed by \citet{lester-etal-2021-power} might hurt the performance for CBQA.

% We suspect that the results of \citet{roberts-etal-2020-much} for T5-Base/Large are better than our baselines (compare with ``no KPs'' column in Table \ref{tab:results:qa}) because we use the checkpoint provided by \citet{lester-etal-2021-power}.

\begin{table}[ht!]
\centering
\begin{tabular}{lcc}
\hline
\textbf{Model} &  \multicolumn{2}{c}{\textbf{TriviaQA}} \\
& \textbf{Dev}
& \textbf{Test} \\
\hline
T5-BASE  + KPs & 24.9 & 28.1  \\
T5-LARGE + KPs & 28.9 & 32.5 \\
\hline
BERT-base \cite{fevry-etal-2020-entities} & 28.9 &  \\
% T5-BASE \cite{roberts-etal-2020-much}  & 23.8 & 29.1   \\
% T5-LARGE \cite{roberts-etal-2020-much} & 28.7 & 35.9  \\
% T5-11B \cite{roberts-etal-2020-much}   & 42.3 & 50.1  \\
EaE \cite{fevry-etal-2020-entities}    & 43.2  & 53.4  \\
T5-11B+SSM \cite{roberts-etal-2020-much} & 51.0  & 60.5 \\
TOME-2 \cite{jong2022mention} & 54.6 & 65.8 \\
\hline
\end{tabular}
\caption{Comparison with other closed-book QA approaches for TriviaQA.}
\label{tab:results:qa_related}
\end{table}

\subsubsection{Fact Checking}
We present experimental results for the fact checking task in Table  \ref{tab:results:fact}.
The use of KPs brings significant improvements for the three model sizes on the FEVER dataset.
Compared to recent works that use LMs with external memory, our T5-Base + KPs model has similar performance to EaE \citep{fevry-etal-2020-entities}, and T5-LARGE + KPs achieves results competitive to TOME-2 \citep{jong2022mention} model.
TOME-2 achieves better results than EaE and T5 + KPs because of the granularity of its memory.
While TOME-2 has an external memory with 150M entries that store fine-grained information about entities, 
both EaE and our model have a memory with about 1M entries only.
Our KP training method allows to increase the granularity of KPs in a straightforward manner.
For instance, one can use multiple KPs per entity, where each KP is trained using a subset of the triples that mention the entity.
We leave the investigation on multiple KPs per entity as a future work.

\begin{table}[h!]
\centering
\begin{tabular}{lcc}
\hline
\textbf{Model} & \multicolumn{2}{c}{\textbf{FEVER}}  \\
& no KPs & KPs \\
\hline
T5-SMALL & 60.2 & \textbf{61.4}  \\
T5-BASE  & 61.3 & \textbf{63.4} \\
T5-LARGE & 63.0 & \textbf{65.2}  \\
\hline
EaE \citep{fevry-etal-2020-entities} & \multicolumn{2}{c}{63.6}   \\
TOME-2 \citep{jong2022mention}      & \multicolumn{2}{c}{68.1}  \\
\hline
\end{tabular}
\caption{Comparing LMs performances on fact checking dataset, FEVER (test split) \citep{thorne-etal-2018-fever}, with and without KPs as a knowledge source, measured by accuracy (\%). Two baselines, Entities-as-Experts (EaE, \citet{fevry-etal-2020-entities}) and MentionMemory (TOME-2, \citet{jong2022mention}), are included.}
\label{tab:results:fact}
\end{table}

\subsubsection{Relation Classification}
Table \ref{tab:results:relation} presents experimental results for the relation classification task using the original TACRED dataset.
Following previous papers, we use F1 as the metric and report results for the test set.
Similar to the other two tasks,
in relation classification KPs also provide performance improvements for all three model sizes.
Interestingly, 
T5-Base + KPs outperform T5-LARGE by almost one point in F1.

\begin{table}[h!]
\centering
\begin{tabular}{lcc}
\hline
\textbf{Model} & \multicolumn{2}{c}{\textbf{TACRED}}  \\
& no KPs & KPs \\
\hline
T5-SMALL & 64.3 & \textbf{66.1} \\
T5-BASE  & 68.3 & \textbf{70.0} \\
T5-LARGE & 69.1 & \textbf{69.8} \\
\hline
EaE \cite{fevry-etal-2020-entities} & \multicolumn{2}{c}{70.2}   \\
KnowBERT \cite{peters-etal-2019-knowbert}      & \multicolumn{2}{c}{71.5}  \\
\hline
\end{tabular}
\caption{Experimental results on relation classification.}
\label{tab:results:relation}
\end{table}

Compared to previous papers that use knowledge augmented approaches,
T5-Base + KPs achieves performance similar to EaE and is competitive with KnowBERT.
It is important to note that KnowBERT uses entity types as input while both EaE and our method do not use that additional information.

\subsection{Ablation Experiments}
\subsubsection{KP$\rightarrow$Encoder vs. KP$\rightarrow$Decoder}
The use of Encoder-Decoder LM gives us the flexibility to introduce KPs at either encoder or decoder.
In all experiments presented so far in the paper, KPs are concatenated to the word embeddings of the input sequence and given as input to the encoder (KP$\rightarrow$Encoder).
However, one can instead concatenate KPs to the output of the encoder, which makes KPs accessible only by the decoder via cross-attention (KP$\rightarrow$Decoder).
In Table \ref{tab:results:kpdec} we present a comparison of the results of KP$\rightarrow$Encoder and KP$\rightarrow$Decoder for the QA datasets.
In both cases, KPs were trained using T5-Base model and we report results for the dev set.
We believe KP$\rightarrow$Encoder achieves better performance because it allows interaction (self-attention) between input and KPs in the encoder, which gives the model more opportunity to select and extract the correct information from KPs.
On the other hand, the advantage of KP$\rightarrow$Decoder is that its training is faster because it is a simpler optimization problem as the error does not have to be back-propagated through the frozen encoder and KPs are used via cross-attention in the decoder only. 
KP$\rightarrow$Decoder requires 3x less training iterations to converge compared to KP$\rightarrow$Enc. 

\begin{table}[h!]
\centering
\begin{tabular}{lcc}
\hline
\textbf{Dataset} & \textbf{KP$\rightarrow$Enc} & \textbf{KP$\rightarrow$Dec} \\
\hline
Simple Questions & \textbf{58.9} & 57.7 \\
Entity Questions & \textbf{33.3} & 32.8 \\
TriviaQA         & \textbf{24.9} & 24.2 \\
\hline
\end{tabular}
\caption{Comparison of models that inject KPs into encoder (KP$\rightarrow$Enc) vs models that inject KPs into decoder (KP$\rightarrow$Dec).}
\label{tab:results:kpdec}
\end{table}

\subsubsection{Entity Linking vs. Searching in KP Space}
Retrieving relevant KPs given an input is a fundamental task that has direct impact on the usefulness of KPs.
Beyond entity linking, 
another approach to retrieve KPs is to transform the input into a single dense vector, then search for the most similar vectors in the KP space.
We have experimented with this strategy by training an external encoder that creates a vector representation of the input.
The external encoder has the same architecture and size of the respective T5 model.
We use the TEKGEN dataset \cite{agarwal-etal-2021-knowledge}, which contains Wikipedia sentences mapped to Wikidata triples, as a source of noisy-labeled data to train the encoder via contrastive loss.
KPs are kept frozen during the training of the input encoder.

Table \ref{tab:results:el_vs_search} presents a comparison between T5 + KP models that use either entity linking or search in the KP space.
The results were computed on TriviaQA.
We can see in the results that entity linking performs better for both model sizes.
We conjecture that searching in KP space does not work well because KPs  are not optimized to be used in search. KPs are trained to memorize knowledge in a way that can later be extracted by the LM.

\begin{table}[h!]
\centering
\begin{tabular}{lcc}
\hline
\textbf{Model} & \textbf{Ent. Linking} & \textbf{KP Search} \\
\hline
T5-SMALL+KPs & \textbf{19.4} & 17.9 \\
T5-BASE+KPs  & \textbf{24.9} & 23.2 \\
\hline
\end{tabular}
\caption{Evaluation of the LM performances on TriviaQA with KPs retrieved with Entity Linking and Searching in KP space.}
\label{tab:results:el_vs_search}
\end{table}

% \subsubsection{Initialization of KPs: word embeddings vs random.}
% - show training curves?
% - with word embeddings: converges faster, easier to train, i.e. less sensitive to hyperparameters
% - show numbers regarding token accuracy in training set?

\section{Conclusions}
We present a new method for training soft prompts that can be used to extend the world knowledge of LMs.
We demonstrate the generality and usefulness of the resulting KPs by employing the same set of KP to improve the LM performance in three different tasks: question answering, fact checking, and relation classification.
Although in this work we focused on the use of KPs for injecting knowledge into LMs,
we believe that entity-centric KPs can be seen as a general purpose knowledge base embedding approach.
We leave the investigation of KPs as a general KB embebedding approach for future work.

\bibliography{anthology,custom}
\bibliographystyle{acl_natbib}

% \appendix

% \section{Example Appendix}
% \label{sec:appendix}

% This is a section in the appendix.

\end{document}